%% file: arxiv_preprint.tex
\documentclass{article}

\usepackage{microtype}
\usepackage{graphicx}
\usepackage{subfigure}
\usepackage{booktabs} 
\usepackage{multirow}
\usepackage[table,xcdraw]{xcolor}

\usepackage{hyperref}

\usepackage[accepted]{arxiv}

\usepackage{amsmath}
\usepackage{amssymb}
\usepackage{mathtools}
\usepackage{amsthm}

\usepackage[capitalize,noabbrev]{cleveref}
\usepackage{hyperref}

\usepackage{verbatim}

\usepackage{xcolor}

\theoremstyle{plain}

\theoremstyle{definition}

\theoremstyle{remark}

\newcommand{\be}{\begin{equation}}
\newcommand{\ee}{\end{equation}}

\newcommand{\ds}{\displaystyle}

\DeclareMathAlphabet{\mathmybb}{U}{bbold}{m}{n}
\newcommand{\1}{\mathmybb{1}}

\let\emptyset\varnothing

\usepackage[textsize=tiny]{todonotes}

\icmltitlerunning{OPSurv: Orthogonal Polynomials Quadrature Algorithm for Survival Analysis}

\begin{document}

\twocolumn[
\icmltitle{OPSurv: Orthogonal Polynomials Quadrature Algorithm for Survival Analysis}



\icmlsetsymbol{equal}{*}

\begin{icmlauthorlist}
\icmlauthor{Lilian W. Bia\l okozowicz}{equal,bai}
\icmlauthor{Hoang M. Le}{equal,bai,york}
\icmlauthor{Tristan Sylvain}{equal,bai}
\icmlauthor{Peter A. I. Forsyth}{bai}
\icmlauthor{Vineel Nagisetty}{bai}
\icmlauthor{Greg Mori}{bai}
\end{icmlauthorlist}
\icmlaffiliation{bai}{Borealis AI, Canada.}
\icmlaffiliation{york}{York University, Canada.}

\icmlcorrespondingauthor{Lilian Bia\l okozowicz}{lilian.wong@borealisai.com}

\icmlkeywords{Orthogonal polynomials approximation, mortgage default, survival analysis}

\vskip 0.3in
]



\printAffiliationsAndNotice{\icmlEqualContribution} 


\begin{abstract}


This paper introduces the Orthogonal Polynomials Quadrature Algorithm for Survival Analysis (OPSurv), a new method providing time-continuous functional outputs for both single and competing risks scenarios in survival analysis. OPSurv utilizes the initial zero condition of the Cumulative Incidence function and a unique decomposition of probability densities using orthogonal polynomials, allowing it to learn functional approximation coefficients for each risk event and construct Cumulative Incidence Function estimates via Gauss--Legendre quadrature. This approach effectively counters overfitting, particularly in competing risks scenarios, enhancing model expressiveness and control. The paper further details empirical validations and theoretical justifications of OPSurv, highlighting its robust performance as an advancement in survival analysis with competing risks.
\end{abstract}

\input{sections/introduction_2}
\input{sections/related_work_2}

\input{sections/method}
\input{sections/experimental_results}
\input{sections/conclusion}

\section*{Acknowledgements}
It is a pleasure to thank Vincent Jeanselme and Dominique Payette for very helpful discussions.

\bibliography{biblio}
\bibliographystyle{icml2024}

\newpage
\appendix
\onecolumn

\input{sections/appendix}

\end{document}

%% file: sections/introduction_2.tex
\section{Introduction}
Survival analysis is a statistical method that focuses on predicting the time until the occurrence of an event of interest, often faced with the challenge of censored observations. Censoring occurs when the event of interest has not happened for certain subjects during the study period, a common scenario in various domains, such as medical research and equipment maintenance.  This paper focuses on the key goal in survival analysis, learning a probability density that captures time to event in the face of these censored observations.

In the realm of survival analysis, a significant portion of existing research operates under the assumption of a single risk factor, such as equipment failure or patient mortality post-transplant~\cite{luck2017deep}. This approach, while useful in certain scenarios, presents a notable limitation as it oversimplifies situations where multiple risk factors may simultaneously influence the time to an event. Recognizing this gap, this paper contributes to the emerging body of work that addresses survival analysis in a competing risks framework. Here, the occurrence of one event can be influenced by multiple, potentially interrelated causes, offering a more nuanced understanding of the factors at play.

A recurrent challenge in survival analysis, exacerbated in the presence of competing risks, is the propensity of models to overfit. This is predominantly due to the high ratio of noise to signal inherent in survival data. Traditional models often combat this through significant constraints on their capacity~\cite{sylvain2021exploring,survivalForestsCompetingRisks}, which can sometimes render the model insufficiently expressive. Our methodology, utilizing the approximation capabilities of Hermite polynomials, presents a novel solution to this issue. By providing a more granular control over the model's expressiveness, our approach effectively mitigates overfitting, thereby enhancing the model's generalisability.

The primary contributions of our research are threefold. Firstly, we introduce the Orthogonal Polynomials Quadrature Algorithm for Survival Analysis (OPSurv), a pioneering approach in survival analysis tailored to settings involving competing risks. This method leverages orthogonal polynomial approximation to finely tune the expressiveness of the model, which outputs time-continuous functions representing survival probabilities and other essential distributions. Secondly, our method demonstrates robust performance across various scenarios, underlining its versatility and efficacy. Finally, we supplement our empirical findings with theoretical justifications, providing a comprehensive validation of our approach. Together, these contributions mark a significant advancement in the field of survival analysis, particularly in addressing the complexities introduced by competing risks.

%% file: sections/related_work_2.tex
\section{Related Work}\label{sec:relatedWork2}

\subsection{Classical Results}
Survival analysis started as time-to-event analysis for a single event. One of the earliest models used in the medical literature was the Kaplan--Meier estimator \cite{kaplan}; the model does not take into account individual attributes. Later came the Cox Proportional Hazards Model (CoxPH) \cite{CoxPH} which estimates the hazard function for each individual. Models in the proportional hazards family assume that the Hazard function $\lambda$ for a client with attribute $x$ is in the form 
\begin{equation}
    \lambda(t|x) = \lambda_0(t) \exp(\beta \cdot x).
    \label{coxPH}
\end{equation} The gist of such models is that $\lambda_0(t)$ is independent of $x$. It is called "proportional" because for two subjects $x_1$ and $x_2$, the ratio of their hazards is constant in time.

The Fine--Gray Model \cite{finegray} extends the traditional CoxPH proportional approach by applying it to modeling the \textit{cumulative distribution function} (CDF) of the survival function instead of the hazard function. However, the resulting hazard function is still in the form \eqref{coxPH}. Inherently, all proportional hazard models are limited by their assumptions on the form of the hazard rates, which may not align well with results from clinical trials \cite{stensrud2019, stensrud2020}. Our proposed method OPSurv does not assume anything about the underlying distribution of the survival function.

\subsection{Machine Learning Approaches}
The development of survival models evolved with the development of machine learning. \cite{faraggi} was among the earliest papers to propose the use of neural networks to survival analysis. Unlike proportional hazard-type models, a feed-forward network was used to learn the relationship between the hazard function and the input covariates.

Random Survival Forests \cite{survivalForests} for single risk was proposed and later was extended to cover competing risks \cite{survivalForestsCompetingRisks}. Random Survival Forests remains one of the most widely applied techniques for its ease of use and explainability. In fact, many single-risk survival models were adopted to the competing risks setting by treating failures from the primary cause of interest as events and failures from other causes as censored observations. Such methods carry the prefix ``cause-specific".

Advances in deep learning also started to influence survival analysis models. DeepSurv \cite{deepsurv} is a Cox proportional hazards feed-forward neural network for modeling interactions between a patient’s covariates and treatment effectiveness. Deep Survival Machines \cite{deepSurvivalMachines} estimates the survival function as a mixture of individual parametric survival distributions.

Other milestones in deep survival analysis are DeepHit \cite{lee2018deephit}, DeSurv \cite{deSurv} and most recently, Neural Fine--Gray \cite{neuralFineGray}. DeepHit learns the joint distribution of survival times and events directly: namely, given the covariates $x$ of a patient, a time window of $T$ (discrete) time steps and $K$ competing risks, the model output is a vector of length $K\cdot T$, $[y_{r, t}]_{r, t}$, where $y_{r, t}$ is the joint probability $P(t,r|x)$ that patient will experience the risk $r$ at time $t$. Given that the CDF for $P(t,r|x)$ is not known, it is estimated by DeepHit as follows
\begin{equation}
F(t, r|x) := P(S \leq t, r| x) \approx \sum_{\tau =0}^{t}  y_{r, \tau}.
\label{DHestF}
\end{equation} For OPSurv, no such approximation is necessary because $F(t,r|x)$ is one of the outputs of the model. In fact, while DeepHit returns a vector with discrete values, OPSurv returns smooth functions of time. Hence, there is no need for extrapolation for OPSurv should the time fall outside of the set of discrete time points.

DeSurv took on a very different approach, pioneering the use of differential equations in the Survival Analysis literature. In particular, it can be credited to be the first in the Survival literature to leverage the fact that if two functions $b(t)$ and $a(t)$ (also known as the \textit{anti-derivative}) are related as follows:
\begin{equation}
    a(t) = \ds \frac{d}{dt}b(t) \text{ and } b(0)=0,
\end{equation} then $b(t)$ can be estimated by means of quadrature
\begin{equation}
    b(t) = \frac{t}{2} \sum_{i=1}^\Gamma w_i a\left( \ds \frac{t}2 (u_i + 1)\right),
    \label{generalQ}
\end{equation}  where $(u_i)_{i=1}^\Gamma$ and $(w_i)_{i=1}^\Gamma$ are the Gauss--Legendre Quadrature nodes and weights of order $\Gamma$ respectively.

DeSurv and OPSurv are different in their outputs and the way quadrature is applied. Like most other existing methods, DeSurv outputs probabilities, while OPSurv generalizes the data to return time-continuous functions. 

Quadrature is used in DeSurv as an immediate step to transform model outputs to $F(t,r|x)$. The neural network outputs positive outputs, $g_r(t|x)$, which are transformed as follows
\begin{equation}
    x,t\xrightarrow[]{\text{NN}} g_{r}(t|x) \xrightarrow[]{\text{qudrature}} g_{r}=\ds \frac{d u}{dt} \xrightarrow[]{\zeta(u)} F(t,r|x),
    \label{DeSurvFlow}
\end{equation} where $\zeta(t)$ can be $\tanh$ or any strictly increasing CDF on the positive x-axis.

For OPSurv, quadrature is used to connect the probability densities $f_r$ and their corresponding CDFs, leveraging a unique property in all Survival problems, that is, CDFs satisfy the initial condition $F|_{t=0}=0$. In the context of \eqref{generalQ}, OPSurv sets $a(t)=f_r(t|x)$ and $b(t)=F_r(t|x)$ and arrives at the quadrature formula \eqref{cifFormula}.

Neural Fine--Gray \cite{neuralFineGray} generalises Sumo-Net \cite{rindt} to competing risks, leveraging monotonic neural networks to model CDFs for each risk. The key is the decomposition
\begin{equation}
F_r(t|x) = P(r|x) \cdot P(T\leq t| r,x).
\label{nfgCIF}
\end{equation} Client attributes are embedded as $E(x)$ and $P(r|x)$ in \eqref{nfgCIF} above is modeled by a multi-layer perceptron with a SoftMax layer $B$ (which is to ensure that $\sum_r F_r(t|x) \leq 1$). Furthermore, $P(T\leq t| r,x)$, which is monotonically increasing, is modeled by a monotone positive network $M_r$ as $1-\exp(-t \times M_r(t, E(x)))$.

%% file: sections/method.tex
\section{Method}

    
    
    

In the method section, we present OPSurv, an algorithm in Survival analysis that provides time-continuous function estimations for survival probability densities, individual and overall \textit{cumulative distribution functions} (CDFs) in scenarios involving single or multiple risks. This approach offers a more general representation of data and includes mechanisms to adjust the degree of approximation, aiding in the prevention of overfitting. A key feature of OPSurv is its use of the initial condition $F|_{t=0}=0$ for all CDFs, enabling the correlation of probability densities with their respective CDFs through Gauss--Legendre quadrature. Differing from parametric models that presuppose specific underlying distributions, OPSurv utilizes a functional analytic perspective. It leverages Orthogonal Polynomials to derive estimation coefficients, effectively describing both probability densities and their CDFs. The upcoming sections detail the specifics of the algorithm and the loss functions employed in our study.




\subsection{Problem Formulation}
OPSurv utilizes a functional-analytic approach to address the Survival analysis problem, focusing on clients and their associated risks. A client $x$ and a risk $e \in [1, \dots, E]$, among $E$ possible risk classes, are considered. The lifetime of a client, denoted as $T$, is a continuous random variable. The Survival Function is defined as $S(t|x)=P(T>t|x)$, where $e=\emptyset$ signifies censored data. For each risk class, the probability density $f_e(t|x)$ corresponds to the conditional survival function $S(t|x, e)$, and $F_e(t|x)$ represents the CDF.

In OPSurv, the densities $f_e(t|x)$ are estimated by decomposing them with respect to a special basis (as detailed in \eqref{densityFormula}). The neural network in this framework learns the coefficients $a_{j,e}(x)$ related to this basis. These coefficients are also employed to derive the corresponding $F_e(t|x)$ through Gauss--Legendre quadrature. OPSurv outputs weights $(\alpha_e(x))_{e=1}^E$, allowing the construction of the \textit{cumulative incidence function} (CIF) as $\alpha_e(x) F_e(t|x)$. 

\begin{figure*}[t]
\vspace{-0.2cm}
\includegraphics[width=.9\textwidth]{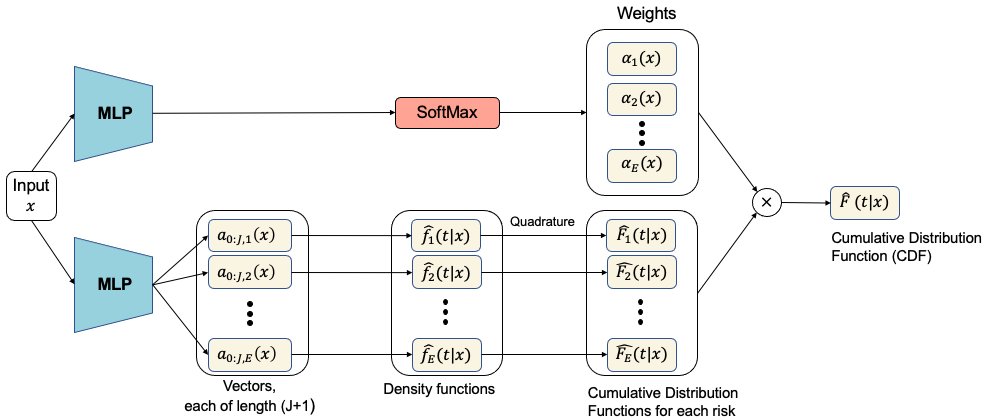}
  \centering
  \vspace{-2.5mm}
  \caption{OPSurv method overview. Client attributes $x$ are passed into 2 neural networks, which output a set of weights and $E$ vectors of length $(J+1)$ respectively. They are coefficients of the probability densities, which can also be used to compute the CDFs and with the weights, the CIFs for each risk.}
  \label{fig:method_overview}
  \vspace{-4mm}
\end{figure*}

The overall methodology of OPSurv, including the interactions between client lifetimes, risk classes, and the neural network's role in estimating survival functions, is depicted in Figure~\ref{fig:method_overview}.

\subsection{OPSurv: the algorithm}

First, to estimate the densities $f_e(t|x)$, we express them as
\be
\hat{f}_e(t|x) := \left(\sum_{j=0}^J a_{j, e}(x) h_j(t) \right)^2 \ds \frac{e^{- t^2}}{W}, 
\label{densityFormula}
\ee where ${h}(t)$ are the normalized (physicists') Hermite polynomials (see Appendix \ref{hermite}) and the denominator $W:=W(a_{j,e}(x))$ are the normalizing weights
\be
W(a_{j,e}(x)) := \ds \sum_{j=0}^J a_{j,e}(x)^2 .
\ee 

An outline of the proof of the decomposition \eqref{densityFormula} will be provided in Section \ref{proof}. Theoretically, the coefficients $a_{j,e}(x)$ would be given by an integral 
\begin{equation}
    a_{j,e}(x) := \int_\mathbb{R} f_e(t|x)^{1/2} h_j(t) e^{-t^2/2} dt,
\label{aEst}
\end{equation} which can be understood as the projection of $f_e(t|x)^{1/2}$ onto the basis $(h_j(t))_{j=0}^\infty$ with respect to the measure $\exp(-t^2/2) dt$. However, the densities $f_e(t|x)$ are unknown to us and the coefficients $a_{j,e}(x)$ will be the outputs of the OPSurv neural network.

Next, to get from $f_e(t|x)$ to $F_e(t|x)$, OPSurv leverages the initial condition in Survival problems satisfied by all CDFs, that is,
\begin{equation}
F_e(0|x) = 0,
\label{initialCondition}
\end{equation} as well as the fact that
\be
f_e(t|x) = \frac{d}{dt} F_e(t|x).
\label{derivRelation}
\ee Equations \eqref{initialCondition} and \eqref{derivRelation} together allow one to use Gauss--Legendre quadrature to estimate $F_e(t|x)$ using the coefficients $\{a_{j,e}(x)\}_{j=0}^J$:
\begin{equation}
F_e(t|x) \approx \hat{F}_e(t|x):=\ds \frac{t}{2} \sum_{\gamma = 1}^{\Gamma} w_\gamma \hat{f}_e \left( \frac{t}{2} (r_\gamma + 1)\right),
\label{cifFormula}
\end{equation} where $\hat{f}_e(t|x)$ is given by \eqref{densityFormula}, and $(w_\gamma, r_\gamma)_{\gamma=1}^{\Gamma}$ are the Gauss--Legendre quadrature weights and nodes respectively of order $\Gamma$\footnote{Note that these quadrature nodes and weights are independent of $f_e(t|x)$ or $F_e(t|x)$. They are computed from the zeros of Legendre polynomial of degree $\Gamma$ and are available in Numpy and SciPy libraries.}.

\subsection{Loss Functions}

In OPSurv, two distinct loss functions are employed: the likelihood loss $\mathcal{L}_{\text{LL}}$ and the ranking loss $\mathcal{L}_{\text{rank}}$. The primary objective of the likelihood loss, $\mathcal{L}_{\text{LL}}$, is to enhance the model's precision in forecasting event probabilities. The ranking loss, $\mathcal{L}_{\text{rank}}$, is on the other hand tailored to address the difficulties posed by the high noise-to-signal ratios often present in medical and mortgage default datasets. It focuses not on the precise prediction of event probabilities but rather on the correct ordering of these probabilities among different examples. This approach is particularly beneficial for tasks such as identifying the most at-risk patients in a medical cohort, where the relative ranking of risk is more critical than the exact probability of event occurrence.

The likelihood loss $\mathcal{L}_{\text{LL}}$, akin to the negative log likelihood, is defined as:
\begin{multline}
    \mathcal{L}_{\text{LL}} := - \sum_{i=1}^N  \left[
    \underbrace{  \1\left\{  e_i \not= \emptyset \right\} \log \left( \alpha_{e_i} \hat{f}_{e_i}(s_i | x_i) \right) }_{\text{if event happened}} \right.
 \\+  \left. \underbrace{ \1\left\{ e_i = \emptyset  \right\}  \log\left( 1 - \sum_{e\in E} \alpha_e(x) \hat{F}_e(s_i|x_i) \right)}_{\text{if censored}} \right] . 
    \label{L1Def}
\end{multline}

The ranking loss $\mathcal{L}_{\text{rank}}$ is formulated over admissible pairs, determined by the \textit{Admissibility Function} $A(m, n) := \1 \left(e_m \neq \emptyset, s_m < s_n \right)$. For a smooth positive convex function $\eta(x, y) := \exp(-(x-y))$, $\mathcal{L}_{\text{rank}}$ is expressed as:
\begin{multline}
\mathcal{L}_{\text{rank}}:= \sum_{m=1}^N \sum_{\substack{n= 1 \\ n\not= m}}^N A(m, n) \\
\cdot \eta \left( \alpha_{e_m}(x_m)\hat{F}_{e_m}(s_m| x_m),
\alpha_{e_m}(x_n) \hat{F}_{e_m}(s_m | x_n) \right)
\label{L2Def}
\end{multline}

The combined loss function used in OPSurv is a weighted sum of these two losses:
\begin{equation}
    \mathcal{L} = \alpha_1 \mathcal{L}_{\text{LL}} + \alpha_2 \mathcal{L}_{\text{rank}},
\end{equation}
where $\alpha_1$ and $\alpha_2$ are scaling factors to balance the different magnitudes of $\mathcal{L}_{\text{LL}}$ and $\mathcal{L}_{\text{rank}}$.


\subsection{Theoretical justification for the approach} \label{proof}
Here is an outline of the proof in order to help the reader have a better grasp of the ideas behind OPSurv. The core of OPSurv lies in the decomposition of the density functions $f_e(t|x)$ as in \eqref{densityFormula}.
The key observation is the following:
\begin{equation}
    \zeta(t) := f_e(t|x)^{\frac 1 2} e^{\frac{t^2}{2}} \in L^2(e^{-t^2}dt).
    \label{zetaDef}
\end{equation} This functional space is special because the set of polynomials is dense in this space (we will illustrate why). Such density is critical for two reasons:
\begin{enumerate}
\item The function, $\zeta(t)$, can be estimated by polynomials (in the $L^2$-sense); and there are coefficients $(a_j)_{j=1}^J$ such that
\begin{equation}
\zeta(t) \approx \zeta_J(t) := \sum_{j=0}^J a_j h_j(t);
\label{zetaEq1}
\end{equation}

\item Moreover, if the polynomial basis are orthogonalized (that can be done by the Gram--Schmidt process with respect to the measure $e^{-t^2} dt$), $\zeta(t)$ can be represented by (physicists') Hermite Polynomials. By the Parseval identity, such orthogonality implies that 
\begin{equation}
\|\zeta_J\|_{L^2}^2
= \ds \int \left(\sum_{j=0}^J a_j h_j(t) \right)^2 e^{-t^2} dt = \sum_{j=0}^J |a_{j}|^2.
\label{parseval}
\end{equation} Without the density of polynomials in this space, \eqref{parseval} may be just an inequality (also known as the Bessel's inequality).
\end{enumerate} 
As a result, ${\zeta_J(t)}/{\|\zeta_J\|_{L^2}}$ has functional norm $1$ and it is a density that approximates $f_e(t|x)^{1/2} e^{t^2/2}$ by \eqref{zetaDef}. This proves the decomposition of $f_e(t|r)$ in \eqref{densityFormula}.

Finally, we go back to prove the density of polynomials in $L^2(e^{-t^2}dt)$ by a classical result in approximation theory, the Riesz's Theorem \cite{riesz}, which states that if a measure $d\nu$ (which is $e^{-t^2}dt$ in our case) satisfies the Moment Condition, that is, for some constant $c>0$,
\begin{equation}
    \int_\mathbb{R} e^{c|t|} d\nu < \infty ,
    \label{moment}
\end{equation} the family of polynomials is dense in $L^2(\nu)$. To see that the measure $e^{-t^2}dt$ satisfies the Moment Condition, note that
\begin{equation}
    \int_\mathbb{R} e^{c|t|} e^{-t^2}dt = 2 \int_{t \geq 0} e^{- (t+c/2)^2} e^{\frac{c^2}4} dt < \infty.
    \label{completeSq}
\end{equation} The equality in \eqref{completeSq} follows from a standard completing square argument. This completes the proof.

%% file: sections/experimental_results.tex
\section{Experimental Results}
\subsection{Baselines}

Our study engages in a comparative analysis with two primary categories of methodologies. The first category includes \textbf{dedicated competing risk methods}, a recent suite of techniques explicitly crafted for the complexities of competing risks scenarios. This category encompasses approaches like DeepHit \citep{lee2018deephit}, DeSurv \cite{deSurv}, and the Neural Fine-Gray model \cite{neuralFineGray}. The second category encompasses \textbf{traditional single risk survival analysis methods}, which we have adapted to our framework by independently modeling each risk. In this category, we compare our approach with established methods such as survival forests \cite{survivalForests}, the cs-Cox Proportional Hazard Model \citep{CoxPH}, and Deep Survival Machines \citep{deepSurvivalMachines}, each representing a cornerstone in the field of survival analysis.

\subsection{Datasets}

Our approach is evaluated using three datasets:

\begin{itemize}
    \item \textbf{Freddie Mac}: The Freddie Mac Single Family Loan-Level Dataset~\cite{FreddieMac2024} includes mortgage attributes, loan origination, and performance data from 1999 to 2023. The primary event of interest is 60-day delinquency, with early-termination as a competing risk. The dataset is divided into three splits based on macroeconomic environments (Table~\ref{tab:FMSplits}), with 375,000 training records, 125,000 for validation, and 100,000 for testing in each split. Event distribution statistics are in Table~\ref{tab:FMSplits}. The observation window spans 48 months, with distinct test periods from the training/validation periods. 29 covariates are used, including loan purpose and interest rates.

    \item \textbf{SEER}\footnote{\url{https://seer.cancer.gov/}}: This dataset contains data on breast cancer patients diagnosed between 1992 and 2017, following the preprocessing steps in Neural Fine--Gray \cite{neuralFineGray}. The primary event is death from breast cancer (BC), with cardiovascular disease (CVD) as a competing risk. It includes 658,354 patients and 24 covariates such as age and tumor size. 

    \item \textbf{METABRIC}: The Molecular Taxonomy of Breast Cancer International Consortium dataset provides gene expression profiles and clinical characteristics from 1,904 breast cancer patients. Data preparation follows the process in~\cite{deepsurv}. METABRIC is the only single-risk dataset in this study.

\end{itemize}

Statistics about the SEER and METABRIC datasets are in Table~\ref{tab:medical-datasets} and Figure~\ref{fig:time_to_event}.

\subsection{Training and evaluation methodology}



This section describes our experimental framework. We employ multilayer perceptron (MLP) networks with 2 hidden layers, each comprising of 32 hidden units using ReLU activation functions. In medical datasets, the data is partitioned into training, validation, and testing splits in the ratio of $64\%/16\%/20\%$. Each experiment is conducted five times, each with a unique random seed affecting both network initialization and data splitting. We report metrics as $\text{mean} (\text{standard deviation})$. Experiments were implemented in the Pytorch framework~\cite{pytorch}, adhering to the PyCox\footnote{\url{https://pypi.org/project/pycox}} structure. For learning-based methods, models were trained over 100 epochs with a batch size of 200 and a learning rate of $1e-3$, using the Adam optimizer~\cite{Kingma2014AdamAM} in its default configuration. Hyperparameters were optimized by evaluating random combinations on the validation sets.



We report results on the following combination of ranking-based and temporal prediction-based metrics:
\begin{enumerate}
\itemsep0em 
\item C-index \cite{concordanceIndex}
\item Time-dependent C-index \cite{timeconcordanceindex} at the $0.25$, $0.5$ and $0.75$ quantiles of the uncensored population event times, specific to the data set.
\item  Time-dependent Brier Score \cite{timeBrier} 
\item Weighted Integrated Brier Score, with inverse-probability-of-censoring weighting (IPCW)\footnote{\url{https://github.com/havakv/pycox/blob/master/pycox/evaluation/ipcw.py}}. We followed DeSurv to use this library, for such weighting compensates for censored data by assigning more weights to uncensored data points with similar characteristics.
\end{enumerate}

\subsection{OPSurv achieves strongly competitive performance in both types of survival problems} \label{quantResults}
Tables~\ref{tab:freddiemac-datasplit-I-paidoff} and \ref{tab:freddiemac-datasplit-I-delinquency} detail results on the Freddie Mac dataset Split I. As we can see, our approach achieves state-of-the-art performance on mortgage default prediction. In addition, when benchmarked on classical survival datasets, our approach is highly competitive against a set of relevant baselines, as outlined in tables ~\ref{tab:seer-event-1-BC}, \ref{tab:seer-event-2-CVD} and \ref{tab:metabric}. The complete set of quantitative results can be found in Appendix \ref{fullStats}. 

It may be worth noting that there did not seem to be a champion model that dominated over other methods on all datasets. In particular, both deep learning and non-deep learning approaches seemed to have their respective strengths and weaknesses. To gain a deeper understanding of the models beyond the statistics, we studied the survival and cumulative incidence functions.

\begin{table}[h]
\caption{Quantitative results for early termination in the Freddie Mac dataset - Data Split I (train: 2013-2016, test: 2018-2021). OPSurv exhibits very strong performance compared to baselines, in particular in terms of IBS.}
\label{tab:freddiemac-datasplit-I-paidoff}
\resizebox{\columnwidth}{!}{%
\begin{tabular}{@{}l|c|c@{}}
\toprule
\textbf{Method} &
  \textbf{td-C Index$\uparrow$} &
  \textbf{Integrated Brier Score$\downarrow$} \\ \midrule
cs-CoxPH                  & 0.551 (0.000) & 0.368 (0.001) \\ \midrule
cs-Random Survival Forest & 0.552 (0.001) & 0.340 (0.003) \\ \midrule
Deep Survival Machines    & 0.535 (0.007) & 0.372 (0.004) \\ \midrule
DeepHit                   & 0.537 (0.003) & 0.319 (0.002) \\ \midrule
DeSurv                    & 0.551 (0.002) & 0.347 (0.006) \\ \midrule
Neural Fine-Gray          & \textbf{0.555 (0.001)} & 0.334 (0.003) \\ \midrule
OPSurv                    & 0.548 (0.016) & \textbf{0.276 (0.011)} \\
\bottomrule
\end{tabular}%
}
\end{table}

\begin{table}[h]
\caption{Quantitative results for 60-day delinquency in the Freddie Mac dataset - Data Split I (train: 2013-2016, test: 2018-2021). OPSurv is in this context highly competitive with the other baselines.}
\label{tab:freddiemac-datasplit-I-delinquency}
\resizebox{\columnwidth}{!}{%
\begin{tabular}{@{}l|c|c@{}}
\toprule
\textbf{Method} &
  \textbf{td-C Index$\uparrow$} &
  \textbf{Integrated Brier Score$\downarrow$} \\ \midrule
cs-CoxPH                  & 0.685 (0.001) & 0.089 (0.004) \\ \midrule
cs-Random Survival Forest & 0.633 (0.003) & 0.087 (0.004) \\ \midrule
Deep Survival Machines    & 0.675 (0.011) & \textbf{0.086 (0.001)} \\ \midrule
DeepHit                   & 0.659 (0.012) & 0.096 (0.000) \\ \midrule
DeSurv                    & \textbf{0.700 (0.004)} & \textbf{0.086 (0.000)} \\ \midrule
Neural Fine-Gray          & 0.682 (0.005) & 0.088 (0.000) \\ \midrule
OPSurv                    & 0.641 (0.011) & \textbf{0.086 (0.000)} \\
\bottomrule
\end{tabular}%
}
\end{table}

\begin{table}[h]
\caption{Quantitative results for the primary event (breast cancer) in the SEER dataset. OPSurv is highly competitive against the baselines in this context.}
\label{tab:seer-event-1-BC}
\resizebox{\columnwidth}{!}{%
\begin{tabular}{@{}l|c|c@{}}
\toprule
\textbf{Method} &
  \textbf{td-C Index$\uparrow$} &
  \textbf{Integrated Brier Score$\downarrow$} \\ \midrule
cs-CoxPH                  & 0.780 (0.001) & 0.134 (0.000) \\ \midrule
cs-Random Survival Forest & 0.833 (0.001) & 0.120 (0.000) \\ \midrule
Deep Survival Machines    & 0.795 (0.002) & 0.138 (0.000) \\ \midrule
DeepHit                   & 0.816 (0.004) & 0.125 (0.002) \\ \midrule
DeSurv                    & 0.829 (0.002) & 0.122 (0.001) \\ \midrule
Neural Fine-Gray          & \textbf{0.835 (0.001)} & \textbf{0.118 (0.001)} \\ \midrule
OPSurv                    & 0.821 (0.009) & 0.123 (0.002) \\
\bottomrule
\end{tabular}
}
\end{table}

\begin{table}[h]
\caption{Quantitative results for the competing risk (cardiovascular disease) in the SEER dataset.}
\label{tab:seer-event-2-CVD}
\resizebox{\columnwidth}{!}{%
\begin{tabular}{@{}l|c|c@{}}
\toprule
\textbf{Method} &
  \textbf{td-C Index$\uparrow$} &
  \textbf{Integrated Brier Score$\downarrow$} \\ \midrule
cs-CoxPH                  & 0.853 (0.001) & \textbf{0.052 (0.000)} \\ \midrule
cs-Random Survival Forest & 0.830 (0.002) & 0.059 (0.000) \\ \midrule
Deep Survival Machines    & 0.849 (0.003) & 0.069 (0.001) \\ \midrule
DeepHit                   & 0.852 (0.002) & 0.057 (0.001) \\ \midrule
DeSurv                    & \textbf{0.854 (0.001)} & 0.056 (0.001) \\ \midrule
Neural Fine-Gray          & 0.852 (0.003) & 0.056 (0.001) \\ \midrule
OPSurv                    & 0.831 (0.013) & 0.061 (0.000) \\
\bottomrule
\end{tabular}%
}
\end{table}

\begin{table}[h]
\caption{Quantitative results for the single risk in METABRIC}
\label{tab:metabric}
\resizebox{\columnwidth}{!}{%
\begin{tabular}{@{}l|c|c @{}}
\toprule
\textbf{Method} &
  \textbf{td-C Index$\uparrow$} &
  \textbf{Integrated Brier Score$\downarrow$} \\ \midrule
cs-CoxPH &
  0.639 (0.019) &
  \textbf{0.150 (0.006)} \\ \midrule
cs-Random Survival Forest &
  0.640 (0.014) &
  0.167 (0.004) \\ \midrule
Deep Survival Machines &
  0.543 (0.048) &
  0.173 (0.008) \\ \midrule
DeepHit &
  0.620 (0.025) &
  0.165 (0.010) \\ \midrule
DeSurv &
  0.651 (0.023) &
  0.152 (0.004) \\ \midrule
Neural Fine-Gray &
  \textbf{0.653 (0.018)} &
  0.151 (0.007) \\ \midrule
OPSurv &
  0.589 (0.036) &
  0.196 (0.010) \\
\bottomrule
\end{tabular}%
}
\end{table}

\begin{figure*}[h]
\vspace{-0.2cm}
\includegraphics[width=.7\textwidth]{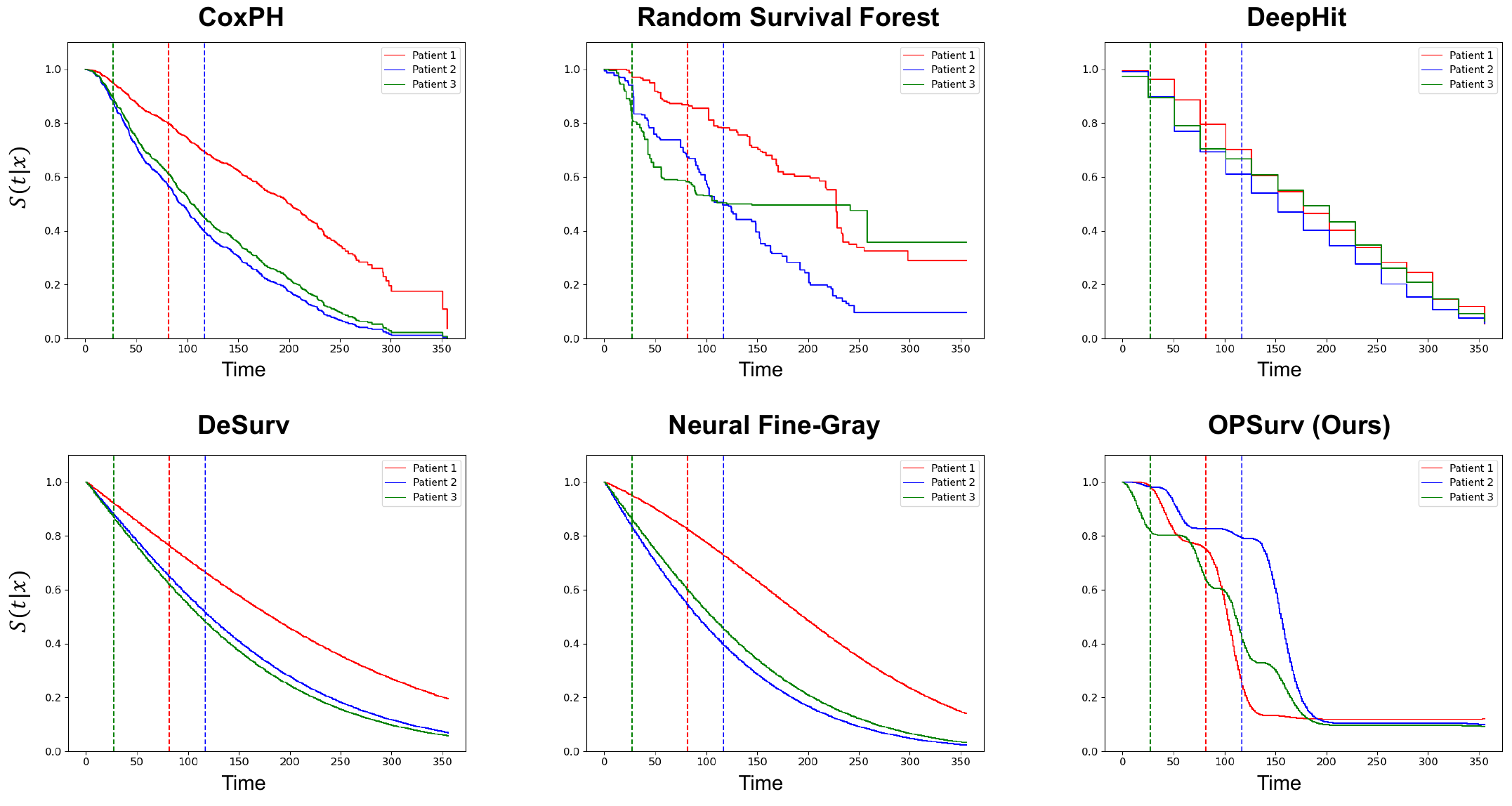}
  \centering
  \vspace{-2.5mm}
  \caption{METABRIC: survival functions of the six models for three random test patients who died. The x-axis is observation time (25 years). Dotted ertical lines indicate the times of death for each patient. Observe that the times of death happened near the inflection points of OPSurv survival functions that preceded a sudden drop.}
  \label{fig:metabric_surv_func}
\end{figure*}

\begin{figure*}[h]
\vspace{-0.2cm}
\includegraphics[width=.8\textwidth]{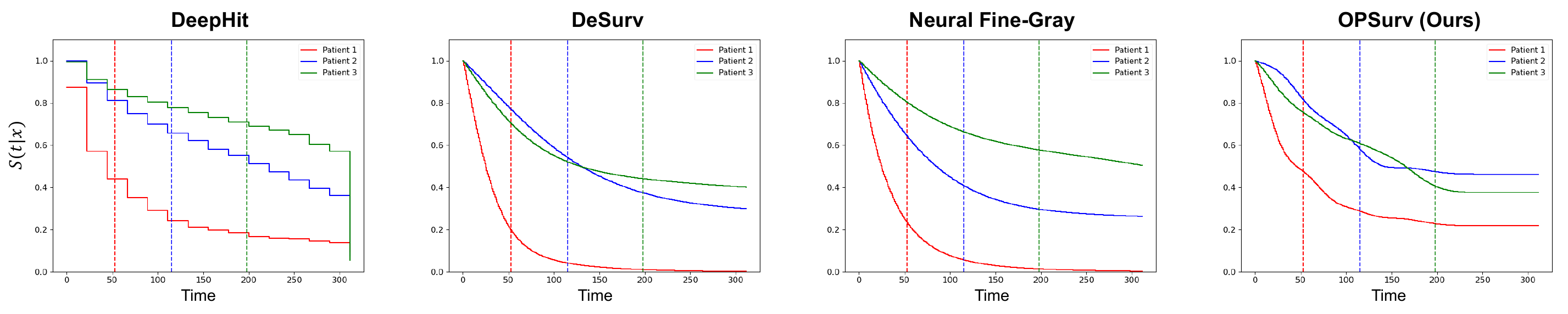}
  \centering
  \vspace{-2.5mm}
  \caption{SEER (primary event): survival functions of four deep learning-based models for three random test patients who died from breast cancer. The x-axis is observation time (25 years). Dotted vertical lines indicate the times of death.}
  \label{fig:seer_surv_event_1}
  \vspace{-4mm}
\end{figure*}

Figures~\ref{fig:metabric_surv_func} and \ref{fig:seer_surv_event_1} show the survival curves of 3 randomly selected patients from the METABRIC and the SEER datasets who died from the primary cause, breast cancer.

\begin{figure*}[h]
\vspace{-0.2cm}
\includegraphics[width=.7\textwidth]{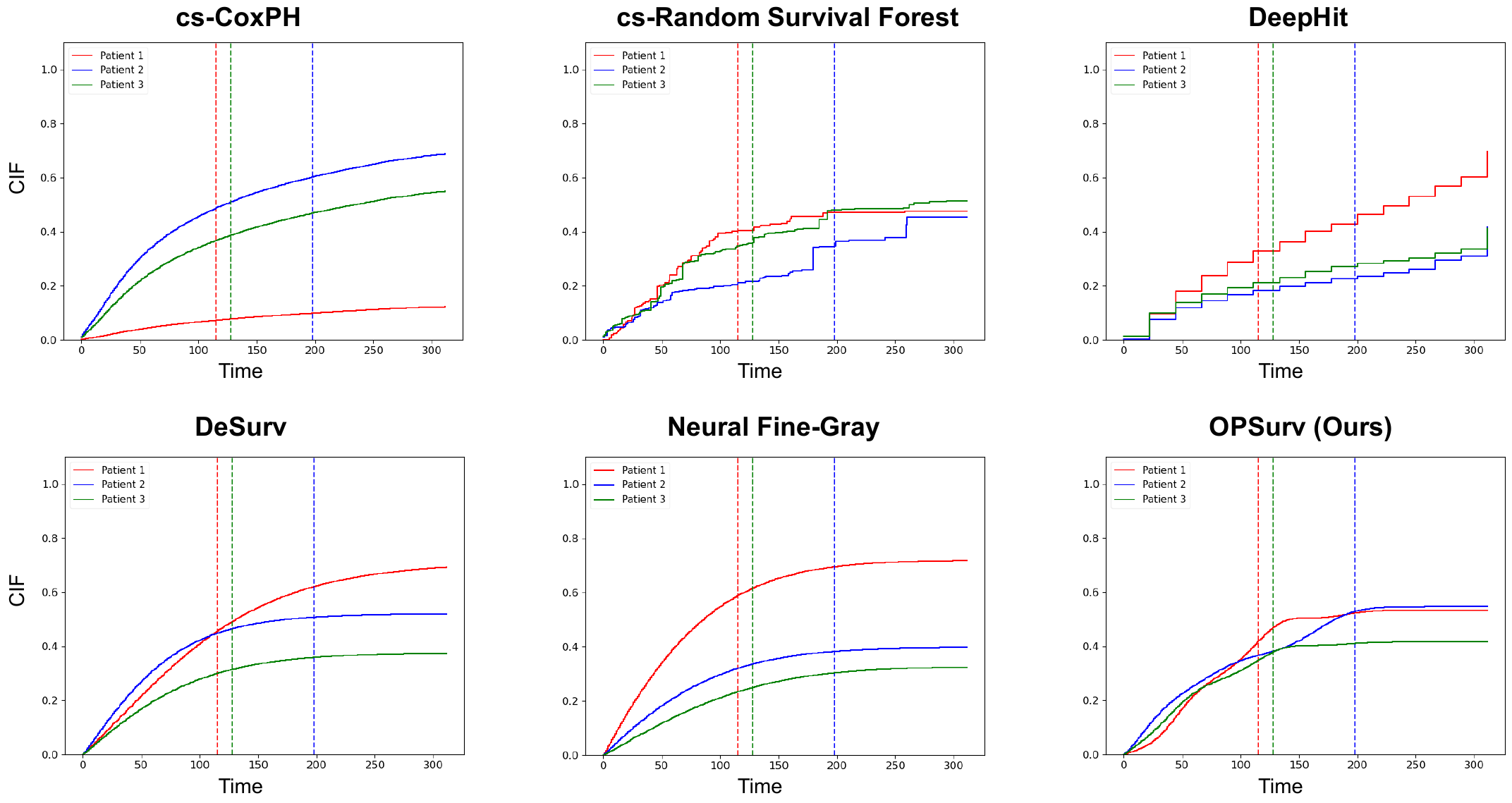}
  \centering
  \vspace{-2.5mm}
  \caption{SEER (primary event): cumulative incidence functions of six models for three random test patients who died from breast cancer. The x-axis is observation time (25 years). Dotted vertical lines indicate the times of death.}
  \label{fig:seer_cif_event_1}
  \vspace{-1mm}
\end{figure*}

\begin{figure*}[h]
\vspace{-0.2cm}
\includegraphics[width=.7\textwidth]{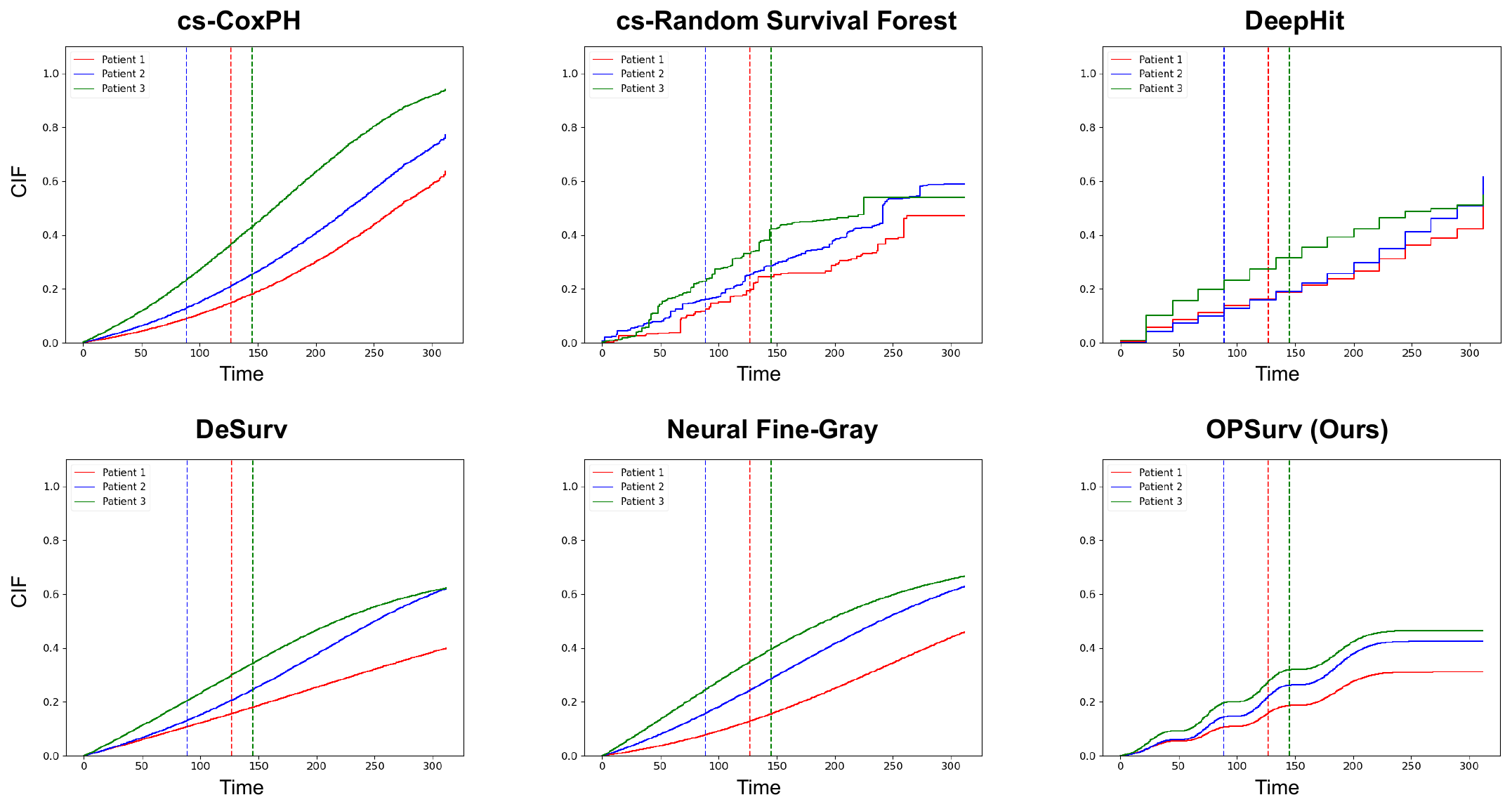}
  \centering
  \vspace{-2.5mm}
  \caption{SEER (competing risk): cumulative incidence functions of six models for three random test patients who died from cardiovascular disease. The x-axis is observation time (25 years). Dotted vertical lines indicate the times of death.}
  \label{fig:seer_cif_event_2}
  \vspace{-1mm}
\end{figure*}

The CIFs for 3 randomly selected patients from the SEER data set who died of the primary cause and competing risks respectively are in figures~\ref{fig:seer_cif_event_1} and \ref{fig:seer_cif_event_2}.

\section{Discussion} 

We observe that in the METABRIC dataset, the times of death coincide with the inflection points of the OPSurv survival curves that precede a steep decline; while in the SEER dataset, the times of death coincide with the elbows of the OPSurv survival curves. Similar occurrences, but less pronounced, can also spotted in some of the DeSurv and Neural Fine--Gray survival curves in figure~\ref{tab:seer-event-1-BC}. These observations may warrant further investigation into the relation between the elbows (or more formally, points of maximum curvature) of the survival functions and the event time, which is now possible for OPSurv, for its survival functions are smooth. OPSurv outputs, being smooth functions, also allow one to compute the Hazard function $\lambda(t)=- d/dt \log S(t)$ directly. To avoid excessive location fluctuations in the survival curves (which are conducive to overfitting), we suggest that the order of approximation, $J$, be chosen to be no more than $15$.

%% file: sections/conclusion.tex
\section{Conclusions}

In this work, we introduce OPSurv, a novel algorithm in survival analysis producing a single set of coefficients capable of describing time-continuous functions, including the density functions $f_r(t|x)$, the cumulative incidence functions $\alpha_r(x) F_r(t|x)$, and the cumulative distribution function $F(t|x)$. The algorithm demonstrates state-of-the-art performance in mortgage default prediction using a large loan repayment dataset and exhibits highly competitive results on various classical competing risks survival datasets.

OPSurv leverages the initial condition $F_r(0|x)=0$ inherent to all Survival problems to establish a connection between the densities and their CDFs via Gauss--Legendre quadrature. Its functional-analytic approach enhances model expressiveness and control while reducing the risk of overfitting. 
The differentiability of the model outputs also comes with advantages. For instance, it allows the direct derivation of the Hazard function $\lambda(t)$ from the equation $\lambda(t) = -d/dt \log S(t)$, as well as other implicit properties of the survival function such as curvature that was mentioned in the discussion. Comparing to existing methods, OPSurv is most versatile for it allows the user to fine-tune the order of approximation to suit the particular dataset, which can mitigate the problem of overfitting.

\section{Impact statement}
This paper presents work of which the goal is to advance the field of Machine Learning. There are many potential societal consequences of our work, none of which we feel must be specifically highlighted here.

%% file: sections/appendix.tex
\section{Hermite Polynomials} \label{hermite}
The Hermite polynomials presented in this paper, $h_j(x)$ are normalized (physicists') Hermite polynomials that satisfy the following orthogonality relations
\be
\int_{\mathbb{R}} h_i(t) h_j(t) e^{-t^2} dt = \delta_{ij}.
\ee In particular, 
\be
h_0(t) = \pi^{-1/4}, h_1(t) = \sqrt{2} \pi^{-1/4} t.
\ee The Hermite polynomials are a special family of orthogonal polynomials which occur naturally in many problems in mathematical physics. The reader may refer to \cite{simon1971distributions} for more details.



\section{Scaling the Input} \label{scalingInput}

\begin{figure*}[h]
\vspace{-0.2cm}
\includegraphics[width=.7\textwidth]{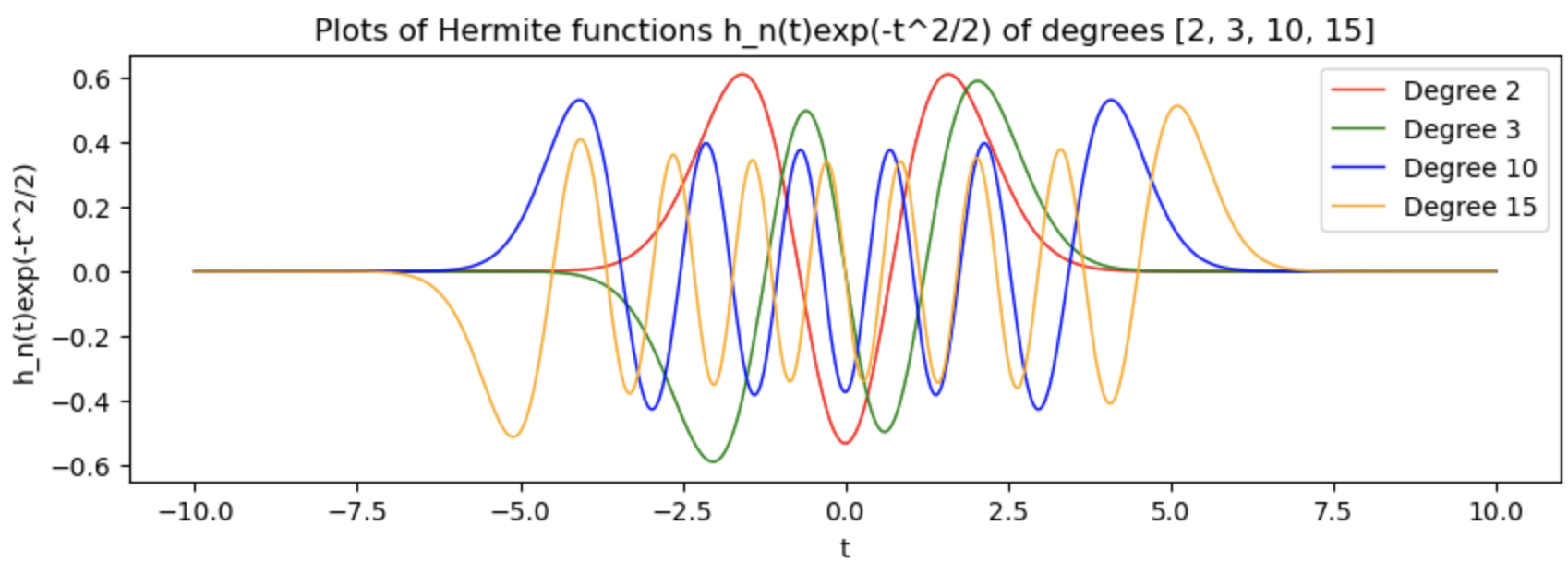}
  \centering
  \vspace{-2.5mm}
  \caption{Plots of Hermite functions of degrees 2, 3, 10 and 15. Observe that they decay exponentially fast outside the interval $[-5, 5]$.}
  \label{fig:hermite_plot}
  \vspace{-4mm}
\end{figure*}

To ensure that precision is not lost in the computation, we suggest scaling the input properly so that it does not fall onto the domain on which the orthogonal functions, $h_n(t)e^{-t^2/2}$, are very close to zero. Our suggestion is to scale as follows: let $T$ be the maximum of all survival or observation times in the data and $r_1 < r_2 <\dots < r_\Gamma$ be the set of quadrature nodes which are symmetric around the $y$-axis, define
\begin{equation}
\tilde{t} := \ds \frac{t}{T} \cdot \frac{10}{\ds r_\Gamma +1},
\end{equation} which achieves the goal that $|\frac{\tilde{t}}{2} (r_\gamma + 1)| \leq 5$. To see why this inequality holds, recall that in \eqref{cifFormula},
\begin{equation}
\hat{F}_e(t|x):=\ds \frac{t}{2} \sum_{\gamma = 1}^{\Gamma} w_\gamma \hat{f}_e \left( \frac{t}{2} (r_\gamma + 1)\right),
\label{recallF}
\end{equation} and $\hat{f}_e$ are composed of Hermite functions of the form $\tilde{h}_n(t) = h_n(t) e^{-t^2/2}$. When $n\leq 20$, $h_n(t)$ decays exponentially fast outside of $[-5, 5]$\footnote{Rigorous asymptotic expressions of Hermite functions have been well studied in the physics literature, see \cite{simon1971distributions} for example.}, as illustrated in Figure \ref{fig:hermite_plot}.

Hence, it is important to ensure that the inputs to $\hat{f}_e$ in \eqref{recallF} sit within the range $[-5, 5]$ or the output of $\hat{f}_e$ might be rounded to $0.0$ during computation. We would also like to emphasize that this scaling procedure is tied to the quadrature order of your choice, $\Gamma$ because the quadrature nodes $r_\gamma$ are related to $\Gamma$.


\section{Model hyperparameters} \label{hyperparams}
Similar to DeSurv, OPSurv employs a uniform MLP for all learning-based methods. For the baseline models, default settings with minimal modifications were used. For DeepHit, we set $T = 15$ as suggested by the authors. For DeSurv, we tried their default settings with $\Gamma = 15$, before increasing $\Gamma$ to $20$. We observed that there was no improvement for DeSurv when increasing the order of quadrature, while OPSurv seemed to have benefited from slightly higher order of quadrature $\Gamma=20$. We set $J=11$ for the METABRIC dataset, and $J=8$ for the SEER and Freddie Mac datasets. 

\section{Dataset Statistics} \label{Appendix:datasets}
In this section, we provide statistics about the three datasets studied in this paper. In particular, we show the distribution of events and censored data in each dataset. It is important to note that these datasets are very different in volume.

\begin{table}[h]
\caption{Data splits for the Freddie Mac Single Family Loan dataset. The data was split into three non-overlapping testing periods. The noticeable differences among the splits could be attributed to the drastically macroeconomic conditions during the three time periods.}
\label{tab:FMSplits}
\vskip 0.15in
\begin{center}
\begin{small}
\begin{sc}
\begin{tabular}{lcccr}
\toprule
\textbf{Data Split} & \textbf{Training/Validation} & \textbf{Testing} & \textbf{Events in Training Data} & \textbf{Events in Testing Data} \\\midrule
\multirow{3}{*}{I} & 2013-2017 & 2018-2021 & Censored: 375.308 (75.06\%) & Censored: 26,566 (26.57\%)\\
                                       & & & Paid Off: 117,137 (23.43\%) & Paid Off: 67.455 (67.46\%)\\
                                       & & & Delinquency: 7,555 (1.51\%) & Delinquency: 5,979 (5.98\%) \\
\midrule
\multirow{3}{*}{II} & 2009-2012 & 2013-2016 & Censored: 254,692 (50.94\%) & Censored: 75,062 (75.62\%) \\
                                        & & & 235,886 (47.18\%) & Paid Off: 23,427 (23.43\%)\\
                                        & & & Delinquency: 9,422 (1.88\%) & Delinquency: 1,511 (1.51\%) \\
\midrule
\multirow{3}{*}{III} & 2004-2007 & 2008-2011 & Censored: 346,177 (69.23\%) & Censored: 39,867 (39.87\%) \\
                                        & &  & Paid Off: 138,736 (27.75\%) & Paid Off: 51,514 (51.54\%) \\
                                     & & & Delinquency: 15,087 (3.02\%) & Delinquency: 8.619 (8.62\%)\\
\bottomrule
\end{tabular}
\end{sc}
\end{small}
\end{center}
\vskip -0.1in
\end{table}


\begin{table*}[h]
\caption{Statistics about the two medical datasets considered in this paper: METABRIC (single risk) and SEER (competing risks). The METABRIC dataset contains a much higher percentage of events than the SEER dataset, but it is also a much smaller dataset than SEER.}
\label{tab:medical-datasets}
\vskip 0.15in
\begin{center}
\begin{small}
\begin{sc}
\begin{tabular}{lccccr}
\toprule
\textbf{Dataset} & \textbf{Type} & \textbf{Records} & \textbf{Features} & \textbf{Events} & \textbf{Censored} \\
\midrule
METABRIC         & Single Risk   & 1,904            & 9                 & 1,103 (57.93\%)     & 801 (42.07\%)     \\
\midrule
\multirow{2}{*}{SEER} & \multirow{2}{*}{Competing} & \multirow{2}{*}{658,354} & \multirow{2}{*}{24} & BC: 108,686 (16.51\%) & \multirow{2}{*}{512,216 (77.80\%)} \\
\cmidrule(lr){5-5}
                 &               &                  &                   & CVD: 37,452(5.69\%) &                   \\\bottomrule
\end{tabular}
\end{sc}
\end{small}
\end{center}
\vskip -0.1in
\end{table*}

\begin{figure}[h]
\vskip 0.2in
\begin{center}
\centerline{
\includegraphics[width=0.45\columnwidth]{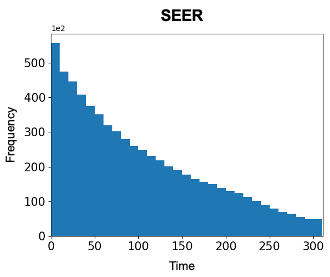}
\includegraphics[width=0.45\columnwidth]{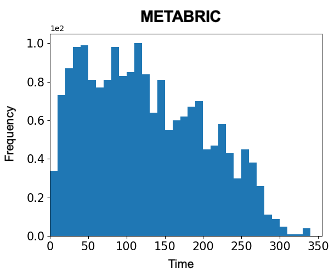}
}
 \caption{Histograms of time-to-event in SEER (left) and METABRIC (right) datasets respectively. Observe that the distribution in SEER follows a more regular pattern than METABRIC, which is much smaller in size but contains a much higher percentage of events.}
\label{fig:time_to_event}
\end{center}
\vskip -0.2in
\end{figure}

\section{Complete performance statistics} \label{fullStats}
We present the full set of performance statistics we obtained from our experiments, including the mean and standard deviation (in parentheses) out of 5 runs. 

\begin{table*}[h]
\caption{Quantitative results for early termination in the Freddie Mac dataset - Split I (train: 2013-2016, test: 2018-2021).}
\label{tab:freddiemac-datasplit-I-paidoff-full}
\resizebox{\textwidth}{!}{%
\begin{tabular}{@{}|l|c|c|c|c|c|c|c|c|@{}}
\toprule
\textbf{Method} &
  \textbf{td-C Index$\uparrow$} &
  \textbf{Integrated Brier Score$\downarrow$} &
  \textbf{td-C Index 25\textsuperscript{th}$\uparrow$} &
  \textbf{td-C Index 50\textsuperscript{th}$\uparrow$} &
  \textbf{td-C Index 75\textsuperscript{th}$\uparrow$} &
  \textbf{Brier Score 25\textsuperscript{th}$\downarrow$} &
  \textbf{Brier Score 50\textsuperscript{th}$\downarrow$} &
  \textbf{Brier Score 75\textsuperscript{th}$\downarrow$} \\ \midrule
cs-CoxPH                  & 0.551   (0.000) & 0.368 (0.001) & 0.560 (0.000) & 0.539 (0.000) & 0.547 (0.000) & 0.154 (0.000) & 0.277 (0.000) & 0.358 (0.000) \\ \midrule
cs-Random Survival Forest & 0.552 (0.001)   & 0.340 (0.003) & 0.532 (0.001) & \textbf{0.554 (0.001)} & 0.558 (0.001) & \textbf{0.152 (0.000)} & 0.283 (0.000) & 0.359 (0.002) \\ \midrule
Deep Survival Machines    & 0.535 (0.007)   & 0.372 (0.004) & \textbf{0.567 (0.006)} & 0.519 (0.005) & 0.529 (0.006) & 0.154 (0.001) & \textbf{0.258 (0.001)} & 0.352 (0.004) \\ \midrule
DeepHit                   & 0.537 (0.003)   & 0.319 (0.002) & 0.461 (0.007) & 0.548 (0.003) & 0.540 (0.003) & \textbf{0.152 (0.001)} & 0.291 (0.001) & \textbf{0.326 (0.003)} \\ \midrule
DeSurv                    & 0.551 (0.002)   & 0.347 (0.006) & 0.554 (0.019) & 0.553 (0.002) & \textbf{0.563 (0.001)} & 0.153 (0.001) & 0.284 (0.003) & 0.372 (0.005) \\ \midrule
Neural Fine-Gray          & \textbf{0.555 (0.001)}   & 0.334 (0.003) & 0.561 (0.004) & 0.551 (0.001) & 0.562 (0.000) & 0.156 (0.000) & 0.292 (0.001) & 0.370 (0.004) \\ \midrule
OPSurv &
  0.548 (0.016) &
  \textbf{0.276 (0.011)} &
  0.537 (0.023) &
  0.552 (0.005) &
  0.558 (0.003) &
  0.156 (0.003) &
  0.301 (0.007) &
  0.360 (0.025) \\ \bottomrule
\end{tabular}
}
\end{table*}

\begin{table*}[h]
\caption{Quantitative results for 60-day delinquency in the Freddie Mac dataset - Data Split I (train: 2013-2016, test: 2018-2021).}
\label{tab:freddiemac-datasplit-I-delinquency-full}
\resizebox{\textwidth}{!}{%
\begin{tabular}{@{}|l|c|c|c|c|c|c|c|c|@{}}
\toprule
\textbf{Method} &
  \textbf{td-C Index$\uparrow$} &
  \textbf{Integrated Brier Score$\downarrow$} &
  \textbf{td-C Index 25\textsuperscript{th}$\uparrow$} &
  \textbf{td-C Index 50\textsuperscript{th}$\uparrow$} &
  \textbf{td-C Index 75\textsuperscript{th}$\uparrow$} &
  \textbf{Brier Score 25\textsuperscript{th}$\downarrow$} &
  \textbf{Brier Score 50\textsuperscript{th}$\downarrow$} &
  \textbf{Brier Score 75\textsuperscript{th}$\downarrow$} \\ \midrule
cs-CoxPH                  & 0.685   (0.001) & 0.089 (0.004) & 0.697 (0.001) & 0.677 (0.001) & 0.683 (0.001) & \textbf{0.017 (0.000)} & 0.035 (0.000) & 0.053 (0.000) \\ \midrule
cs-Random Survival Forest & 0.633 (0.003)   & 0.087 (0.004) & 0.638 (0.005) & 0.641 (0.005) & 0.648 (0.002) & \textbf{0.017 (0.000)} & 0.035 (0.000) & 0.053 (0.000) \\ \midrule
Deep Survival Machines    & 0.675 (0.011)   & 0.086 (0.001) & 0.689 (0.013) & 0.667 (0.010) & 0.673 (0.010) & \textbf{0.017 (0.000)} & \textbf{0.034 (0.000)} & 0.053 (0.000) \\ \midrule
DeepHit                   & 0.659 (0.012)   & 0.096 (0.000) & 0.674 (0.010) & 0.682 (0.013) & 0.671 (0.014) & \textbf{0.017 (0.000)} & 0.035 (0.000) & \textbf{0.052 (0.000)} \\ \midrule
DeSurv                    & \textbf{0.700 (0.004)}   & \textbf{0.086 (0.000)} & \textbf{0.709 (0.007)} & \textbf{0.697 (0.002)} & \textbf{0.699 (0.003)} & \textbf{0.017 (0.000)} & 0.035 (0.000) & 0.054 (0.000) \\ \midrule
Neural Fine-Gray          & 0.682 (0.005)   & 0.088 (0.000) & 0.693 (0.006) & 0.674 (0.009) & 0.679 (0.007) & \textbf{0.017 (0.000)} & 0.035 (0.000) & 0.055 (0.000) \\ \midrule
OPSurv &
  0.641 (0.011) &
  \textbf{0.086 (0.000)} &
  0.659 (0.017) &
  0.664 (0.017) &
  0.666 (0.017) &
  0.017 (0.000) &
  0.035 (0.000) &
  0.054 (0.000) \\ \bottomrule
\end{tabular}%
}
\end{table*}

\begin{table*}[h]
\caption{Quantitative results for early termination in the Freddie Mac dataset - Data Split II (train: 2009-2012, test: 2013-2016).}
\label{tab:freddiemac-datasplit-II-paidoff-full}
\resizebox{\textwidth}{!}{%
\begin{tabular}{@{}|l|c|c|c|c|c|c|c|c|@{}}
\toprule
\textbf{Method} &
  \textbf{td-C Index$\uparrow$} &
  \textbf{Integrated Brier Score$\downarrow$} &
  \textbf{td-C Index 25\textsuperscript{th}$\uparrow$} &
  \textbf{td-C Index 50\textsuperscript{th}$\uparrow$} &
  \textbf{td-C Index 75\textsuperscript{th}$\uparrow$} &
  \textbf{Brier Score 25\textsuperscript{th}$\downarrow$} &
  \textbf{Brier Score 50\textsuperscript{th}$\downarrow$} &
  \textbf{Brier Score 75\textsuperscript{th}$\downarrow$} \\ \midrule
cs-CoxPH &
  0.651   (0.000) &
  0.156 (0.000) &
  0.670 (0.000) &
  0.629 (0.000) &
  0.627 (0.000) &
  \textbf{0.067 (0.000)} &
  \textbf{0.118 (0.000)} &
  0.145 (0.000) \\ \midrule
cs-Random Survival Forest &
  0.616 (0.003) &
  0.170 (0.000) &
  0.614 (0.004) &
  0.615 (0.002) &
  0.625 (0.002) &
  0.070 (0.000) &
  0.128 (0.000) &
  0.158 (0.000) \\ \midrule
Deep Survival Machines &
  0.663 (0.005) &
  \textbf{0.155 (0.001)} &
  0.700 (0.007) &
  0.621 (0.006) &
  0.610 (0.008) &
  0.070 (0.001) &
  0.121 (0.001) &
  0.146 (0.002) \\ \midrule
DeepHit &
  0.657 (0.010) &
  0.156 (0.006) &
  0.674 (0.023) &
  0.634 (0.013) &
  0.622 (0.016) &
  0.068 (0.001) &
  0.123 (0.003) &
  \textbf{0.144 (0.005)} \\ \midrule
DeSurv &
  \textbf{0.677 (0.002)} &
  0.160 (0.002) &
  0.696 (0.009) &
  \textbf{0.644 (0.002)} &
  \textbf{0.636 (0.002)} &
  \textbf{0.067 (0.000)} &
  0.120 (0.001) &
  0.148 (0.001) \\ \midrule
Neural Fine-Gray &
  0.676 (0.005) &
  0.161 (0.002) &
  \textbf{0.707 (0.004)} &
  0.643 (0.006) &
  0.635 (0.007) &
  0.068 (0.000) &
  0.121 (0.001) &
  0.149 (0.001) \\ \midrule
OPSurv &
  0.626 (0.029) &
  0.191 (0.016) &
  0.651 (0.034) &
  0.623 (0.027) &
  0.624 (0.022) &
  \textbf{0.067 (0.000)} &
  0.119 (0.003) &
  0.148 (0.008) \\ \bottomrule
\end{tabular}%
}
\end{table*}

\begin{table*}[h]
\caption{ Quantitative results for 60-day delinquency in the Freddie Mac dataset - Data Split II (train: 2009-2012, test: 2013-2016) - }
\label{tab:freddiemac-datasplit-II-delinquency-full}
\resizebox{\textwidth}{!}{%
\begin{tabular}{@{}|l|c|c|c|c|c|c|c|c|@{}}
\toprule
\textbf{Method} &
  \textbf{td-C Index$\uparrow$} &
  \textbf{Integrated Brier Score$\downarrow$} &
  \textbf{td-C Index 25\textsuperscript{th}$\uparrow$} &
  \textbf{td-C Index 50\textsuperscript{th}$\uparrow$} &
  \textbf{td-C Index 75\textsuperscript{th}$\uparrow$} &
  \textbf{Brier Score 25\textsuperscript{th}$\downarrow$} &
  \textbf{Brier Score 50\textsuperscript{th}$\downarrow$} &
  \textbf{Brier Score 75\textsuperscript{th}$\downarrow$} \\ \midrule
cs-CoxPH &
  \textbf{0.817   (0.000)} &
  \textbf{0.021 (0.001)} &
  \textbf{0.824 (0.001)} &
  \textbf{0.811 (0.000)} &
  \textbf{0.799 (0.000)} &
  \textbf{0.005 (0.000)} &
  \textbf{0.011 (0.000)} &
  0.016 (0.000) \\ \midrule
cs-Random Survival Forest &
  0.753 (0.012) &
  0.022 (0.001) &
  0.746 (0.021) &
  0.765 (0.012) &
  0.778 (0.008) &
  \textbf{0.005 (0.000)} &
  \textbf{0.011 (0.000)} &
  0.017 (0.000) \\ \midrule
Deep Survival Machines &
  0.808 (0.009) &
  0.022 (0.000) &
  0.823 (0.009) &
  0.795 (0.013) &
  0.781 (0.014) &
  \textbf{0.005 (0.000)} &
  \textbf{0.011 (0.000)} &
  \textbf{0.015 (0.000)} \\ \midrule
DeepHit &
  0.694 (0.028) &
  0.036 (0.002) &
  0.753 (0.010) &
  0.756 (0.038) &
  0.743 (0.041) &
  \textbf{0.005 (0.000)} &
  \textbf{0.011 (0.000)} &
  \textbf{0.015 (0.000)} \\ \midrule
DeSurv &
  0.807 (0.006) &
  0.022 (0.000) &
  0.820 (0.003) &
  0.796 (0.007) &
  0.783 (0.007) &
  \textbf{0.005 (0.000)} &
  \textbf{0.011 (0.000)} &
  \textbf{0.015 (0.000)} \\ \midrule
Neural Fine-Gray &
  0.765 (0.120) &
  0.022 (0.001) &
  0.768 (0.131) &
  0.772 (0.102) &
  0.764 (0.097) &
  \textbf{0.005 (0.000)} &
  \textbf{0.011 (0.000)} &
  \textbf{0.015 (0.000)} \\ \midrule
OPSurv &
  0.768 (0.022) &
  0.022 (0.000) &
  0.776 (0.037) &
  0.787 (0.028) &
  0.786 (0.027) &
  \textbf{0.005 (0.000)} &
  \textbf{0.011 (0.000)} &
  \textbf{0.015 (0.000)} \\ \midrule
\end{tabular}%
}
\end{table*}

\begin{table*}[h]
\caption{ Quantitative results for early termination in the Freddie Mac dataset - Data Split III (train: 2004-2007, test: 2008-2011).}
\label{tab:freddiemac-datasplit-III-paidoff-full}
\resizebox{\textwidth}{!}{%
\begin{tabular}{@{}|l|c|c|c|c|c|c|c|c|@{}}
\toprule
\textbf{Method} &
  \textbf{td-C Index$\uparrow$} &
  \textbf{Integrated Brier Score$\downarrow$} &
  \textbf{td-C Index 25\textsuperscript{th}$\uparrow$} &
  \textbf{td-C Index 50\textsuperscript{th}$\uparrow$} &
  \textbf{td-C Index 75\textsuperscript{th}$\uparrow$} &
  \textbf{Brier Score 25\textsuperscript{th}$\downarrow$} &
  \textbf{Brier Score 50\textsuperscript{th}$\downarrow$} &
  \textbf{Brier Score 75\textsuperscript{th}$\downarrow$} \\ \midrule
cs-CoxPH &
  0.469 (0.001) &
  0.332 (0.000) &
  0.490 (0.001) &
  0.459 (0.001) &
  0.451 (0.001) &
  0.146 (0.000) &
  0.235 (0.000) &
  0.310 (0.000) \\ \midrule
cs-Random Survival Forest &
  0.492 (0.001) &
  0.330 (0.002) &
  0.473 (0.006) &
  0.490 (0.003) &
  0.426 (0.002) &
  0.147 (0.000) &
  0.238 (0.001) &
  0.313 (0.001) \\ \midrule
Deep Survival Machines &
  0.475 (0.008) &
  0.344 (0.007) &
  0.503 (0.007) &
  0.462 (0.008) &
  0.452 (0.008) &
  \textbf{0.142 (0.001)} &
  \textbf{0.232 (0.002)} &
  0.323 (0.005) \\ \midrule
DeepHit &
  0.491 (0.006) &
  0.333 (0.007) &
  0.491 (0.007) &
  0.482 (0.010) &
  \textbf{0.491 (0.008)} &
  0.147 (0.000) &
  0.252 (0.001) &
  0.336 (0.003) \\ \midrule
DeSurv &
  0.474 (0.006) &
  0.338 (0.003) &
  0.494 (0.007) &
  0.478 (0.007) &
  0.444 (0.010) &
  0.148 (0.000) &
  0.244 (0.001) &
  0.321 (0.003) \\ \midrule
Neural Fine-Gray &
  0.471 (0.018) &
  0.338 (0.002) &
  0.491 (0.011) &
  0.470 (0.023) &
  0.430 (0.007) &
  0.149 (0.001) &
  0.247 (0.002) &
  0.327 (0.005) \\ \midrule
OPSurv &
  \textbf{0.499 (0.022)} &
  \textbf{0.269 (0.003)} &
  \textbf{0.523 (0.012)} &
  \textbf{0.512 (0.025)} &
  0.441 (0.008) &
  0.147 (0.001) &
  0.239 (0.003) &
  \textbf{0.308 (0.006)} \\ \bottomrule
\end{tabular}%
}
\end{table*}

\begin{table*}[h]
\caption{ Quantitative results for 60-day delinquency in the Freddie Mac dataset - Data Split III (train: 2004-2007, test: 2008-2011).}
\label{tab:freddiemac-datasplit-III-delinquency-full}
\resizebox{\textwidth}{!}{%
\begin{tabular}{@{}|l|c|c|c|c|c|c|c|c|@{}}
\toprule
\textbf{Method} &
  \textbf{td-C Index$\uparrow$} &
  \textbf{Integrated Brier Score$\downarrow$} &
  \textbf{td-C Index 25\textsuperscript{th}$\uparrow$} &
  \textbf{td-C Index 50\textsuperscript{th}$\uparrow$} &
  \textbf{td-C Index 75\textsuperscript{th}$\uparrow$} &
  \textbf{Brier Score 25\textsuperscript{th}$\downarrow$} &
  \textbf{Brier Score 50\textsuperscript{th}$\downarrow$} &
  \textbf{Brier Score 75\textsuperscript{th}$\downarrow$} \\ \midrule
cs-CoxPH &
  0.728   (0.004) &
  0.097 (0.001) &
  0.738 (0.004) &
  0.712 (0.006) &
  0.714 (0.006) &
  \textbf{0.028 (0.000)} &
  0.054 (0.000) &
  0.077 (0.000) \\ \midrule
cs-Random Survival Forest &
  0.715 (0.003) &
  \textbf{0.096 (0.001)} &
  0.706 (0.005) &
  0.719 (0.004) &
  0.727 (0.002) &
  \textbf{0.028 (0.000)} &
  \textbf{0.053 (0.000)} &
  \textbf{0.075 (0.000)} \\ \midrule
Deep Survival Machines &
  0.739 (0.013) &
  0.101 (0.001) &
  0.746 (0.010) &
  0.729 (0.016) &
  0.730 (0.016) &
  \textbf{0.028 (0.000)} &
  \textbf{0.053 (0.000)} &
  0.077 (0.001) \\ \midrule
DeepHit &
  0.722 (0.014) &
  0.108 (0.002) &
  0.737 (0.014) &
  0.744 (0.014) &
  \textbf{0.747 (0.011)} &
  \textbf{0.028 (0.000)} &
  0.054 (0.000) &
  0.078 (0.001) \\ \midrule
DeSurv &
  0.741 (0.005) &
  0.098 (0.001) &
  0.749 (0.004) &
  0.731 (0.005) &
  0.729 (0.006) &
  \textbf{0.028 (0.000)} &
  \textbf{0.053 (0.000)} &
  0.077 (0.000) \\ \midrule
Neural Fine-Gray &
  \textbf{0.749 (0.004)} &
  0.101 (0.001) &
  \textbf{0.756 (0.004)} &
  \textbf{0.738 (0.003)} &
  0.739 (0.004) &
  0.029 (0.000) &
  0.055 (0.000) &
  0.081 (0.000) \\ \midrule
OPSurv &
  0.704 (0.013) &
  0.099 (0.001) &
  0.703 (0.008) &
  0.717 (0.022) &
  0.709 (0.016) &
  0.029 (0.000) &
  0.054 (0.000) &
  0.078 (0.000) \\ \bottomrule
\end{tabular}%
}
\end{table*}

\begin{table*}[h]
\caption{Quantitative results for breast cancer (primary event) on the SEER dataset.}
\label{tab:seer-event-1-BC-full}
\resizebox{\textwidth}{!}{%
\begin{tabular}{@{}|l|c|c|c|c|c|c|c|c|@{}}
\toprule
\textbf{Method} &
  \textbf{td-C Index$\uparrow$} &
  \textbf{Integrated Brier Score$\downarrow$} &
  \textbf{td-C Index 25\textsuperscript{th}$\uparrow$} &
  \textbf{td-C Index 50\textsuperscript{th}$\uparrow$} &
  \textbf{td-C Index 75\textsuperscript{th}$\uparrow$} &
  \textbf{Brier Score 25\textsuperscript{th}$\downarrow$} &
  \textbf{Brier Score 50\textsuperscript{th}$\downarrow$} &
  \textbf{Brier Score 75\textsuperscript{th}$\downarrow$} \\ \midrule
cs-CoxPH                  & 0.780 (0.001) & 0.134 (0.000) & 0.858 (0.001) & 0.816 (0.001) & 0.773 (0.002) & 0.042 (0.000) & 0.079 (0.000) & 0.116 (0.000) \\ \midrule
cs-Random Survival Forest & 0.833 (0.001) & 0.120 (0.000) & \textbf{0.906 (0.001)} & \textbf{0.870 (0.001)} & \textbf{0.830 (0.002)} & \textbf{0.035 (0.000)} & \textbf{0.065 (0.000)} & \textbf{0.098 (0.001)} \\ \midrule
Deep Survival Machines    & 0.795 (0.002) & 0.138 (0.000) & 0.878 (0.002) & 0.832 (0.002) & 0.789 (0.002) & 0.041 (0.000) & 0.080 (0.000) & 0.122 (0.000) \\ \midrule
DeepHit                   & 0.816 (0.004) & 0.125 (0.002) & 0.901 (0.001) & 0.859 (0.003) & 0.800 (0.009) & 0.039 (0.000) & 0.076 (0.000) & 0.112 (0.001) \\ \midrule
DeSurv                    & 0.829 (0.002) & 0.122 (0.001) & 0.899 (0.001) & 0.864 (0.002) & 0.825 (0.002) & 0.037 (0.000) & 0.068 (0.000) & 0.101 (0.001) \\ \midrule
Neural Fine-Gray          & \textbf{0.835 (0.001)} & \textbf{0.118 (0.001)} & 0.903 (0.001) & 0.869 (0.001) & \textbf{0.830 (0.001)} & 0.037 (0.000) & 0.067 (0.000) & 0.099 (0.001) \\ \midrule
OPSurv &
  0.821 (0.009) &
  0.123 (0.002) &
  0.894 (0.008) &
  0.855 (0.006) &
  0.808 (0.005) &
  0.039 (0.001) &
  0.071 (0.002) &
  0.104 (0.002) \\
  \bottomrule
  \end{tabular}
  }
\end{table*}

\begin{table*}[h]
\caption{Quantitative results for cardiovascular disease (competing risk) on the SEER dataset.}
\label{tab:seer-event-2-CVD-full}
\resizebox{\textwidth}{!}{%
\begin{tabular}{@{}|l|c|c|c|c|c|c|c|c|@{}}
\toprule
\textbf{Method} &
  \textbf{td-C Index$\uparrow$} &
  \textbf{Integrated Brier Score$\downarrow$} &
  \textbf{td-C Index 25\textsuperscript{th}$\uparrow$} &
  \textbf{td-C Index 50\textsuperscript{th}$\uparrow$} &
  \textbf{td-C Index 75\textsuperscript{th}$\uparrow$} &
  \textbf{Brier Score 25\textsuperscript{th}$\downarrow$} &
  \textbf{Brier Score 50\textsuperscript{th}$\downarrow$} &
  \textbf{Brier Score 75\textsuperscript{th}$\downarrow$} \\ \midrule
cs-CoxPH                  & 0.853   (0.001) & \textbf{0.052 (0.000)} & 0.854 (0.006) & 0.854 (0.003) & 0.854 (0.002) & \textbf{0.009 (0.000)} & \textbf{0.019 (0.000)} & \textbf{0.038 (0.000)} \\ \midrule
cs-Random Survival Forest & 0.830 (0.002)   & 0.059 (0.000) & 0.838 (0.005) & 0.844 (0.004) & 0.844 (0.003) & \textbf{0.009 (0.000)} & \textbf{0.019 (0.000)} & \textbf{0.038 (0.000)} \\ \midrule
Deep Survival Machines    & 0.849 (0.003)   & 0.069 (0.001) & 0.867 (0.005) & 0.859 (0.002) & 0.851 (0.003) & \textbf{0.009 (0.000)} & \textbf{0.019 (0.000)} & 0.040 (0.000) \\ \midrule
DeepHit                   & 0.852 (0.002)   & 0.057 (0.001) & \textbf{0.868 (0.003)} & \textbf{0.863 (0.002)} & \textbf{0.857 (0.002)} & \textbf{0.009 (0.000)} & \textbf{0.019 (0.000)} & \textbf{0.038 (0.000)} \\ \midrule
DeSurv                    & \textbf{0.854 (0.001)}   & 0.056 (0.001) & 0.857 (0.008) & 0.855 (0.003) & 0.853 (0.002) & \textbf{0.009 (0.000)} & \textbf{0.019 (0.001)} & 0.039 (0.000) \\ \midrule
Neural Fine-Gray          & 0.852 (0.003)   & 0.056 (0.001) & 0.852 (0.008) & 0.852 (0.005) & 0.852 (0.003) & \textbf{0.009 (0.000)} & \textbf{0.019 (0.000)} & 0.039 (0.000) \\ \midrule
OPSurv &
  0.831 (0.013) &
  0.061 (0.000) &
  0.832 (0.024) &
  0.830 (0.014) &
  0.831 (0.007) &
  \textbf{0.009 (0.000)} &
  \textbf{0.019 (0.001)} &
  0.040 (0.001) \\ \bottomrule
\end{tabular}%
}
\end{table*}

\begin{table*}[h]
\caption{Quantitative results for breast cancer (single risk) on the METABRIC dataset.}
\label{tab:metabric-full}
\resizebox{\textwidth}{!}{%
\begin{tabular}{@{}|l|c|c|c|c|c|c|c|c|@{}}
\toprule
\textbf{Method} &
  \textbf{td-C Index$\uparrow$} &
  \textbf{Integrated Brier Score$\downarrow$} &
  \textbf{td-C Index 25\textsuperscript{th}$\uparrow$} &
  \textbf{td-C Index 50\textsuperscript{th}$\uparrow$} &
  \textbf{td-C Index 75\textsuperscript{th}$\uparrow$} &
  \textbf{Brier Score 25\textsuperscript{th}$\downarrow$} &
  \textbf{Brier Score 50\textsuperscript{th}$\downarrow$} &
  \textbf{Brier Score 75\textsuperscript{th}$\downarrow$} \\ \midrule
cs-CoxPH &
  0.639 (0.019) &
  \textbf{0.150 (0.006)} &
  0.633 (0.038) &
  0.629 (0.025) &
  0.641 (0.023) &
  0.122 (0.002) &
  0.196 (0.006) &
  \textbf{0.220 (0.016)} \\ \midrule
cs-Random Survival Forest &
  0.640 (0.014) &
  0.167 (0.004) &
  \textbf{0.702 (0.040)} &
  0.662 (0.019) &
  0.635 (0.023) &
  \textbf{0.118 (0.004)} &
  \textbf{0.193 (0.005)} &
  0.231 (0.017) \\ \midrule
Deep Survival Machines &
  0.543 (0.048) &
  0.173 (0.008) &
  0.567 (0.039) &
  0.559 (0.052) &
  0.568 (0.055) &
  0.126 (0.001) &
  0.208 (0.003) &
  0.245 (0.008) \\ \midrule
DeepHit &
  0.620 (0.025) &
  0.165 (0.010) &
  \textbf{0.702 (0.042)} &
  \textbf{0.667 (0.021)} &
  0.606 (0.026) &
  0.122 (0.001) &
  0.199 (0.002) &
  0.239 (0.011) \\ \midrule
DeSurv &
  0.651 (0.023) &
  0.152 (0.004) &
  0.658 (0.057) &
  0.646 (0.037) &
  0.648 (0.023) &
  0.121 (0.003) &
  0.194 (0.005) &
  0.223 (0.012) \\ \midrule
Neural Fine-Gray &
  \textbf{0.653 (0.018)} &
  0.151 (0.007) &
  0.673 (0.043) &
  0.658 (0.020) &
  \textbf{0.657 (0.022)} &
  0.119 (0.003) &
  \textbf{0.193 (0.007)} &
  \textbf{0.220 (0.016)} \\ \midrule
OPSurv &
  0.589 (0.036) &
  0.196 (0.010) &
  0.622 (0.054) &
  0.594 (0.042) &
  0.555 (0.037) &
  0.124 (0.004) &
  0.213 (0.006) &
  0.285 (0.025) \\ \bottomrule
\end{tabular}%
}
\end{table*}
